\documentclass[11pt]{article}

\usepackage[final]{acl}

\usepackage{times}
\usepackage{latexsym}
\usepackage{booktabs}
\usepackage{multirow}
\usepackage[table]{xcolor}
\usepackage{adjustbox}
\usepackage[T1]{fontenc}
\usepackage[edges]{forest}
\usepackage[utf8]{inputenc}

\usepackage{microtype}

\usepackage{inconsolata}
\usepackage{amsmath}
\usepackage{amssymb}

\usepackage{graphicx}

\title{A Survey of Process Reward Models: From Outcome Signals to Process Supervisions for Large Language Models}

\author{
  Congmin Zheng\thanks{Equal Contribution}\textsuperscript{\rm 1}, Jiachen Zhu$^{\ast}$\textsuperscript{\rm 1}, Zhuoying Ou$^{\ast}$\textsuperscript{\rm 1}, Yuxiang Chen\textsuperscript{\rm 2}, \\ \textbf{Kangning Zhang}\textsuperscript{\rm 1},\textbf{Rong Shan}\textsuperscript{\rm 1}\textbf{Zeyu Zheng}\textsuperscript{\rm 3}, \textbf{Mengyue Yang}\textsuperscript{\rm 4},\\ \textbf{Jianghao Lin}\textsuperscript{\rm 1}\thanks{Corresponding author}, \textbf{Yong Yu}\textsuperscript{\rm 1}, \textbf{Weinan Zhang}\textsuperscript{\rm 1}$^\dagger $\\
  \textsuperscript{\rm 1}Shanghai Jiao Tong University, \textsuperscript{\rm 2}University College London, \\ 
  \textsuperscript{\rm 3}Carnegie Mellon University,
  \textsuperscript{\rm 4}University of Bristol\\
    \texttt{\{desp.zcm,gebro13,zoeouzy23,linjianghao,wnzhang\}@sjtu.edu.cn, }
    }

\begin{document}

\maketitle

\begin{abstract}
Although Large Language Models (LLMs) exhibit advanced reasoning ability, conventional alignment remains largely dominated by outcome reward models (ORMs) that judge only final answers. \emph{Process Reward Models} (PRMs) address this gap by evaluating and guiding reasoning at the step or trajectory level. This survey provides a systematic overview of PRMs through the full loop: how to \emph{generate process data}, \emph{build PRMs}, and \emph{use PRMs} for test-time scaling and reinforcement learning. We summarize applications across math, code, text, multimodal reasoning, robotics, and agents, and review emerging benchmarks. Our goal is to clarify design spaces, reveal open challenges, and guide future research toward fine-grained, robust reasoning alignment. To support these efforts, we accompany this survey with an actively updated GitHub repository (\url{https://github.com/despzcm/Survey-of-Process-Reward-Model}).
\end{abstract}

\section{Introduction}

The advent of Large Language Models (LLMs) has reshaped alignment for reasoning~\cite{shao2024deepseekmath,jaech2024openai,yang2025qwen3,bai2025qwen2,he2025skywork},
shifting attention from \emph{outcome-only} supervision to \emph{process-aware} evaluation.
Early pipelines predominantly relied on outcome reward models (ORMs)~\citep{lightman2023let} that judge only final
answers, providing a single coarse signal for long chains of thought. As reasoning tasks grow
longer and more complex, this static, outcome-centric view struggles to capture stepwise
progress, diagnose intermediate errors, or allocate computation adaptively.

\begin{figure}[t]
\centering
\includegraphics[width=\linewidth]{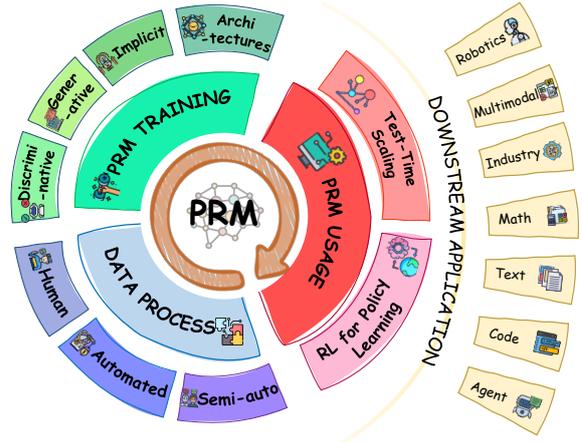}
\caption{The Process Reward Model (PRM) loop that iteratively
\emph{generates data}, \emph{trains PRMs}, and \emph{uses PRMs} to improve policies and produce new data.}
\label{fig:prm_evolution}
\end{figure}

To address this gap, the community has begun to move beyond coarse outcome supervision toward process reward models (PRMs), which explicitly assess and guide reasoning at the step or trajectory level.
As shown in Figure~\ref{fig:prm_evolution},  Process Reward Models coupled with a closed loop:
\emph{generate process data} $\rightarrow$ \emph{train PRMs} $\rightarrow$ \emph{use PRMs}
(test-time scaling or RL) $\rightarrow$ \emph{produce better data}.
This loop transforms reward modeling from a one-shot verdict to an iterative controller of
reasoning, enabling finer credit assignment, richer diagnostics, and improved robustness.

The emergence of PRMs marks a pivotal shift. Rather than relying on single-turn or
rule-based evaluation, PRMs assess partial solutions and trajectories, leverage context for
adaptive “reason-then-rate” verification, and integrate with inference-time controllers and
reinforcement learning (RL) objectives. In this paradigm, supervision becomes proactive:
it not only evaluates but also \emph{steers} search, reflection, and policy updates across
diverse sources of evidence (e.g., retrieved knowledge, programs, or multimodal inputs).

Given these rapid advances, we present a systematic survey of PRMs across the full loop:
\textbf{how to generate data}, \textbf{how to build PRMs}, and \textbf{how to use PRMs}.
Current discussions mainly focus on either test-time scaling paradigms~\cite{zhang2025survey_tts}, broad reward modeling taxonomies~\cite{zhong2025survey_reward_models}, or generic deep RL reward design~\cite{yu2025survey_rl_reward_models}, whereas our PRM survey uniquely targets step-level process reward modeling by organizing the full loop of data generation, PRM building, and usage (test-time scaling and PRM-guided RL) for fine-grained reasoning supervision.

Specifically, this paper is structured as follows.
Sec.~\ref{sec: data process} (\textbf{How to Generate Data}) categorizes process supervision
into \emph{human annotation}, \emph{automated supervision}, and \emph{semi-automated pipelines},
highlighting fidelity–scalability trade-offs.
Sec.~\ref{sec:build-prm} (\textbf{How to Build PRMs}) reviews modeling paradigms, including
\emph{discriminative} vs.\ \emph{generative} objectives, \emph{explicit} vs.\ \emph{implicit}
supervision, and architectural innovations.
Sec.~\ref{sec: usage} (\textbf{How to Use PRMs}) discusses \emph{test-time scaling}
(re-ranking, verification-guided decoding, search) and \emph{PRM-guided RL} (dense stepwise
rewards and credit assignment). Sec.~\ref{sec:application} includes \textbf{applications} spanning math, code, multimodal
reasoning, agents, and high-stakes domains, and Sec.~\ref{sec:benchmark} summarizes \textbf{benchmarks}. Further \textbf{discussions} are provided in Sec.~\ref{sec:discussion}.

\section{How to Generate Data}
\label{sec: data process}
In this section, we address the question of "how to generate data" for training process reward models (PRMs) and categorize existing approaches into three main paradigms: (1) human annotation, (2) automated supervision, and (3) hybrid methods that combine both. Each paradigm reflects a different trade-off between fidelity and scalability, and recent work often integrates multiple strategies to leverage the strengths of one source while mitigating the weaknesses of another.

\subsection{Human Annotation}
\label{sec: Human Annotation}
The earliest and most straightforward form of process supervision comes from direct human annotation, where annotators explicitly verify the correctness of intermediate reasoning steps. PRM800K~\citep{lightman2023let} is a representative example, in which human labelers carefully validated each step of multi-hop reasoning chains. This dataset demonstrated that explicitly capturing human judgments about process correctness can substantially improve PRM training, leading to better alignment and more interpretable reasoning outcomes.

Although resource-intensive and limited in scale, human-curated process data has proven to be a critical foundation: it provides high-fidelity signals, establishes benchmarks for other data generation pipelines, and often serves as seed material to guide more scalable methods.

\subsection{Automated Supervision}
\label{sec:Automated Supervision}
To overcome the bottlenecks of manual labeling, a large body of research explores fully automated approaches that generate process supervision through symbolic verification, consistency checks, execution feedback, or synthetic self-evolution.

Math-Shepherd~\citep{wang2023math} introduced an automated verification pipeline where mathematical reasoning steps are validated using symbolic tools and consistency-checking heuristics, enabling large-scale process supervision without human annotations. FOVER~\citep{kamoi2025fover} uses formal verification tools (e.g., Z3, Isabelle) to automatically generate PRM training data with accurate step-level error labels. OmegaPRM~\citep{luo2024improve} extends this paradigm by using a divide-and-conquer style Monte Carlo Tree Search (MCTS) algorithm to efficiently identify the first error in a reasoning chain, providing a scalable alternative to human judgment. URSA~\citep{luo2025ursa} further advances this line by synthesizing process-level supervision for multimodal mathematical reasoning through a fully automated dual-view pipeline, which employs MCTS-based error localization and misinterpretation insertion engines to construct large-scale process annotations.

Expanding beyond mathematics, MT-RewardTree~\citep{feng2025mt} adapts the MCTS-driven framework to machine translation, leveraging approximate MCTS to generate token-level preference pairs entirely through automatic evaluation and filtering, thereby enabling scalable and fine-grained reward modeling without human annotation. Similarly, CodePRM~\citep{li2025codeprm} employs automated tree search and execution feedback to derive step-level supervision for code reasoning, achieving fully automatic label generation without human involvement. Search-in-Context~\citep{chen2025search} introduces Monte Carlo Tree Search with dynamic retrieval, which automatically constructs intermediate reasoning steps without requiring human-annotated reasoning chains or task-specific rewards.

Some approaches take automation even further. In AlphaMath~\citep{chen2024alphamath}, researchers propose an even more radical approach: deriving pseudo-process supervision directly from outcome supervision, thereby eliminating the need for stepwise labels altogether. More structured methods have also been developed, such as Tree-PLV~\citep{he2024advancing}, which learns preferences over trees of reasoning trajectories automatically constructed via a best-first search algorithm. Building on this trend, rStar-Math~\citep{guan2025rstar} and Qwen2.5-Math PRM~\citep{zhang2025lessons} adopt self-evolutionary and consensus-filtering strategies respectively to create massive reasoning datasets, while EpicPRM~\citep{sun2025efficient} focuses on balancing precision and scale in constructing process-supervised training data.

To improve robustness, SCAN~\citep{ding2025scan} introduces a self-denoising annotation framework that automatically detects and corrects noisy labels, and ~\citet{wang2025towards} proposes a data augmentation strategy based on node merging in the tree structure.

Collectively, these works showcase the promise of automated pipelines: they enable unprecedented scale and efficiency, though they must carefully address error propagation, verifier limitations, and potential misalignment with human reasoning preferences.

\subsection{Semi-automated Approaches}
\label{sec:Semi-automated Approaches}
Between these two extremes, a growing number of works adopt semi-automated approaches, blending selective human input with scalable automated expansion. In multimodal reasoning, this pattern is especially pronounced: VRPRM~\citep{chen2025vrprm} and Athena~\citep{wang2025athena} both construct PRM datasets by starting with limited human-curated reasoning steps and then expanding them with automated verification or synthetic generation, significantly improving data efficiency. ViLBench~\citep{tu2025vilbench} and VisualPRM~\citep{wang2025visualprm} adopt similar strategies in vision-language reasoning, mixing curated samples with large-scale synthetic data to create comprehensive benchmarks.

In more specialized domains,  $\text{MedS}^3$~\citep{jiang2025meds} adopts a self-evolved “slow thinking” paradigm for medical reasoning: it starts from around 8,000 human-curated examples and then automatically expands them via MCTS-based exploration and rule-verifiable trajectory generation, greatly reducing manual workload while retaining domain reliability. Beyond single-domain settings, VersaPRM~\citep{zeng2025versaprm} generates synthetic reasoning data across multiple domains primarily via auto-labeling, with a small-scale manual evaluation conducted to verify the quality of the auto-labeled data.

Practical task-oriented applications also rely on hybrid pipelines. Web-Shepherd~\citep{chae2025web} supervises web navigation reasoning traces by mixing human oversight with automatic checks, while GUI-Shepherd~\citep{chen2025gui} builds the PRM dataset via a dual-pipeline strategy combining diverse trajectories with hybrid human-GPT annotations. Finally, ActPRM~\citep{duan2025efficient} exemplifies active learning in PRM training, selectively querying human annotators only when automated signals are uncertain, thereby reducing labeling costs without sacrificing supervision quality.

These hybrid methods illustrate that carefully combining human anchors with automated pipelines not only mitigates the weaknesses of each approach but also opens up broader applications in domains where neither purely human nor purely automated supervision is sufficient.

\section{How to Build PRMs}
\label{sec:build-prm}

In this section, we answer the question of "how to build PRMs" and categorize PRM training works into four classes: Discriminative PRMs, Generative PRMs, Implicit PRMs, and Other Architectures. Furthermore, we provide detailed discussions of representative methods in each category.

\subsection{Discriminative PRMs.}\label{sec:disc-prm}
A \emph{discriminative} PRM learns a scoring function over intermediate reasoning states to predict per-step correctness, plausibility, or progress. Given an input $x$ and a partial solution $s_{1:t}$, the model outputs a scalar score as Eq.~\ref{eq:pointwise} shows.

\begin{equation}
\label{eq:pointwise}
r_t = \sigma(f_\theta(x, s_{1:t})) \in (0,1)
\end{equation}

\textbf{Pointwise loss.}  
The score $r_t$ can be trained with standard pointwise objectives. Here $\sigma$ is the sigmoid function, and $f_\theta$ denotes the discriminative PRMs. With binary labels $y_t\in\{0,1\}$ or soft labels $y_t\in[0,1]$, one typically uses either binary cross-entropy (BCE) or mean squared error (MSE):
\begin{equation}
 \mathcal{L}_{\text{point}}^{\text{BCE}} = \mathbb{E}\!\big[-y_t\log r_t - (1-y_t)\log(1-r_t)\big],
\end{equation}

\begin{equation}
 \mathcal{L}_{\text{point}}^{\text{MSE}} = \mathbb{E}\!\big[(r_t - y_t)^2\big].
\label{eq:disc-point}
\end{equation}

\textbf{Pairwise (preference) loss.}  
Alternatively, discriminative PRMs can be trained on \emph{relative} preferences between two candidate steps or partial traces $u$ and $v$. The model predicts the probability that $u$ is preferred to $v$:
\begin{equation}
\mathbb{P}_\theta(u \succ v) = 
\sigma\!\big(f_\theta(u)-f_\theta(v)\big),
\label{eq:pair-prob}
\end{equation}
and minimizes a pairwise (preference) loss such as:
\begin{equation}
\mathcal{L}_{\text{pair}} = 
\mathbb{E}\!\big[-\log \mathbb{P}_\theta(u \succ v)\big],
\label{eq:pair-loss}
\end{equation}
which is analogous to the Direct Preference Optimization (DPO) objective used in RLHF.

Discriminative PRMs, viewed as the foundational training paradigm in the history of process-level reward models, have inspired lots of works. DreamPRM~\citep{cao2025dreamprm} alternately trains the PRM and domain weights through a bi-level strategy to generalize across multimodal tasks; PQM~\citep{PQM} recasts PRM as a Q-value ranking problem, aligning rewards by relative ordering; ER-PRM~\citep{zhang2024entropy} injects entropy regularization into the reward objective to avoid overconfident predictions and improve calibration; EDU-PRM~\citep{cao2025bangbuckprocessreward} uses entropy-based uncertainty sampling and weighting to focus training on ambiguous or difficult reasoning steps; Q-RM~\citep{Q-RM} introduces token-level discriminative loss to provide finer-grained feedback on intermediate tokens; BiPRM~\citep{zhang2025bidirectional} seamlessly integrates a parallel right-to-left (R2L) evaluation stream with the conventional L2R flow, allowing later reasoning steps to real-time assist in assessing earlier ones;R-PRM~\citep{R-PRM} designs a loss function that favors logical and structural consistency across reasoning steps; BiRM~\citep{chen2025better} not only evaluates the correctness of previous steps, but also models the probability of future success; CoLD~\citep{zheng2025cold} uses counterfactual guidance to mitigate length bias in reward scoring; and ProgRM~\cite{zhang2025progrm} defines dynamic “progress rewards” that proportionally align process rewards with the degree of task completion.

\subsection{Generative PRMs.}\label{sec:gen-prm}
A \emph{generative} PRM operates in two stages: it first generates a verification or critique chain \(z_t\) (“think”), and then judges or scores the original reasoning step based on that chain (“judge”). Concretely, one can write:

\begin{equation}
\begin{aligned}
    z_t &\sim p_\phi\bigl(z_t \mid x, s_{1:t}\bigr) \\
    r_t &= h_\psi\bigl(x, s_{1:t}, z_t\bigr),
\end{aligned}
\end{equation}

% \begin{equation}
%     r_t &= h_\psi\bigl(x, s_{1:t}, z_t\bigr),
% \end{equation}

where \(p_\phi\) is the generative verifier or critic model, and \(h_\psi\) is a scoring head that maps the generated chain and the step history to a step-level reward \(r_t\). A plausible joint training objective combines a likelihood loss for the verification chain and a supervision term for the step-level reward:

\begin{equation}
\mathcal{L}_{\text{gen}} \;=\; 
-\log p_\phi\bigl(z_t^\star \mid x, s_{1:t}\bigr) \;+\; \lambda \,\mathrm{BCE}\bigl(r_t, y_t\bigr),
\label{eq:gen}
\end{equation}

where \(z_t^\star\) is a reference (e.g., \ human or oracle) critique chain, and \(y_t\) is the ground-truth (or soft) label for the step.

In many works, $h_\psi$ is simply the confidence of the answer logits. Assume token indices \(k_{\mathrm{yes}}\) and \(k_{\mathrm{no}}\) correspond to “yes” and “no” respectively. Then define $r_t$ as the softmax score:
\begin{equation}
r_t  = \frac{\exp(q_{k_{\mathrm{yes}}})}{\exp(q_{k_{\mathrm{yes}}}) + \exp(q_{k_{\mathrm{no}}})}.
\end{equation}

This generative PRM paradigm helps the reward model maintain long reasoning chains (i.e., extended “thinking”) and better understand the semantics of the input.
ThinkPRM~\citep{lee2025rethinking} uses an internal “thinking” loop to simulate generative reflection and enable dynamic reasoning. GenRM~\citep{GenRM} introduces chain-of-thought at inference and uses voting to pick the highest-scoring reasoning chain to improve consistency. GenPRM~\citep{zhao2025genprm} applies generative computation scaling at test time to boost the stability of reward predictions. GRAM-R²~\citep{GRAM-R} self-trains a generative foundation reward model that evolves its own reasoning and reward logic. Process-based Self-Rewarding Language Models~\citep{zhang2025process} allow the model to both generate and assess its own reasoning chains, closing the loop between reasoning and reward. Test-Time Scaling with Reflective Generative Model~\citep{wang2025test} expands inference-time generative capacity and applies reflection to refine reward prediction. GM-PRM~\citep{zhang2025gm} is the first multimodal generative PRM, supporting chain generation in multimodal mathematical reasoning tasks. rStar-Math~\citep{guan2025rstar} strengthens smaller models’ reasoning by evolving deep thinking through self-evolution in its internal reasoning architecture. 

\subsection{Implicit PRMs}\label{subsec:exp-vs-imp}
\label{subsec:exp-vs-imp}

The above discriminative and generative PRM methods all rely on explicit supervision signals derived from annotated reasoning steps; in contrast, implicit PRMs aim to infer fine-grained rewards without step-level labels, by leveraging weaker or indirect supervision such as outcome feedback, model self-evaluation, or consistency constraints. Implicit PRM extracts step rewards from unlabeled trajectories; FreePRM~\citep{sun2025freeprm} trains a reward model without ground-truth process labels by pseudo-labeling via outcome correctness; Self-PRM~\citep{Self-PRM} shows that LLMs under RL training can internally induce a PRM-style self-rewarding capability; SP-PRM~\citep{SP-PRM} transfers reasoning knowledge from an outcome reward model (ORM) into process reward modeling to reduce label dependency; SPARE~\citep{rizvi2025spare} uses one-shot reference guidance to automatically generate supervision signals for intermediate steps; Universal PRM (AURORA)~\citep{tan2025aurora} employs ensemble prompting and reverse verification to produce domain-agnostic self-supervised reward signals; and Process-based Self-Rewarding Language Models let the model generate and evaluate its own reasoning chain, closing the loop for self-supervision.

\subsection{Other Architectural Innovations}\label{sec:other-arch}
Other architectures in the PRM landscape emphasize innovations in model structure, reasoning representations, or system frameworks rather than new loss functions or supervision schemes. For example, GraphPRM~\cite{peng2025rewarding} casts reasoning as a graph of steps and learns structured dependencies among them; ASPRM (AdaptiveStep)~\citep{liu2025adaptivestep} dynamically adjusts the granularity of reasoning steps based on model confidence; Reward-SQL~\citep{zhang2025reward} builds a structured process reward model tailored to the Text-to-SQL domain; RetrievalPRM~\citep{zhu2025retrieval} integrates external retrieval to ground reward predictions and improve cross‐task generalization; OpenPRM~\citep{zhang2025openprm} organizes reward judgments into an open preference tree, supporting branching and domain flexibility; MM-PRM~\cite{du2025mm} provides a unified multimodal PRM architecture and open implementation; Multilingual PRM~\citep{wang2025demystifying} addresses cross‐language CoT transfer through representational mapping across languages; PathFinder-PRM~\cite{pala2025errortypingsmarterrewards} employs a hierarchical error‐aware architecture to distinguish and reward different types of reasoning errors; and Hierarchical Reward Model (HRM)~\citep{wang2025hierarchicalmultisteprewardmodels} proposes layered reward structures aligned with multi‐level reasoning abstractions.

\section{How to Use PRMs}
\label{sec: usage}
In this section, we discuss how to use PRMs and organize their usage into two main paradigms: Test-Time Scaling and Reinforcement Learning for Policy Learning. We further provide detailed discussions of representative methods and developments within each paradigm, highlighting how PRMs guide inference, search, and policy learning through fine-grained step-level feedback.
\subsection{Test-Time Scaling}
\label{sec:tts}
Test-time scaling aims to improve model performance not by enlarging model size but by strategically allocating computation during inference—via candidate sampling, re-ranking, or guided search. PRMs are central to this process, providing fine-grained evaluation of intermediate reasoning steps and trajectories to guide test-time computation.

Early work used PRMs primarily as re-rankers. Studies such as ~\citet{lightman2023let,wang2023math,wang2025visualprm,wang2025athena,zheng2025cold} showed that Best-of-N re-ranking with PRM scores consistently improves final performance, validating PRMs as reliable test-time evaluators.
Building on this foundation, PRMs evolved into generative verifiers. GenPRM~\citep{zhao2025genprm} introduced verification-by-generation, producing reasoning or code checks before scoring candidates. ThinkPRM~\citep{snell2024scaling} fine-tunes long chain-of-thought verifiers with limited process-level labels, enhancing scaling under Best-of-N and beam search. ~\citet{kim2025scaling} formalized reasoning-oriented evaluation as a mechanism for allocating test-time compute more effectively, positioning PRMs as flexible controllers of inference resources.

Parallel efforts integrated PRMs into search and decoding algorithms. PRM-BAS~\citep{hu2025prm} embedded PRMs into beam annealing search, pruning low-quality candidates to improve efficiency. CodePRM~\citep{li2025codeprm} implemented a Generate–Verify–Refine pipeline, using PRMs to detect and correct faulty intermediate code steps. Web-Shepherd~\citep{chae2025web} filtered web-agent trajectories, while other approaches combined PRMs with MCTS or retrieval-augmented reasoning~\citep{chan2025boosting,ma2025static,chen2025search}. Safety-aware scaling was addressed by SAFFRON-1, which reduced costly PRM calls and introduced caching mechanisms to ensure robust, efficient inference under adversarial conditions.

Finally, refinements targeted step-level granularity and adaptivity. AdaptiveStep~\citep{liu2025adaptivestep} dynamically partitions reasoning into finer steps based on confidence, producing sharper PRM judgments. SP-PRM~\citep{xie2025outcomes} extended reward-guided search strategies across multiple granularity levels, from tokens to full responses, enhancing both precision and flexibility.

Together, these developments trace a clear trajectory: from static PRM-based re-ranking, through generative verification and search integration, to adaptive step-level refinements and safety-aware scaling, transforming PRMs into dynamic, scalable controllers of inference.

\subsection{RL for Policy Learning}
\label{sec: RL}
The use of process reward models (PRMs) within reinforcement learning (RL) has become a promising direction for aligning language models with fine-grained reasoning quality. Traditional RL relies on outcome-only supervision, which is sparse and often misaligned with intermediate reasoning steps. By contrast, PRMs provide dense step-level or trajectory-level feedback that can be integrated into RL training loops, offering more stable credit assignment and faster policy learning.

Early explorations established that PRMs could directly replace sparse correctness-based signals with fine-grained supervision during RL. Math-Shepherd~\citep{wang2023math} trained an automatic verifier that scores each intermediate step in math reasoning and used those scores as rewards for PPO, allowing the policy to learn from abundant intermediate feedback when final answers are rare. In a similar vein, ~\citet{dai2024process} demonstrated how line-level PRM signals could be injected into RL training, overcoming the limitations of outcome-only feedback from unit tests and enabling policies to improve across long coding trajectories. Extending this idea to practical domains, Reward-SQL~\citep{zhang2025reward} integrated stepwise PRMs into an online RL loop, showing that process-level signals are especially valuable in text-to-SQL generation, while ReasonRAG~\citep{zhang2025process} applied PRM-guided RL to retrieval-augmented generation agents. Together, these works show that PRMs can serve as actionable dense rewards that significantly improve RL training across reasoning-heavy tasks.

Building on this foundation, several studies refined the formulation of PRM signals within RL objectives. PAV~\citep{setlur2024rewarding} reframed step-level PRM outputs as advantage-like progress indicators, providing dense step-level rewards for RL training of policy models. ER-PRM~\citep{zhang2024entropy} introduced an entropy-regularized framework that embeds PRM rewards into KL-constrained RL objectives, stabilizing training while preserving exploration. PURE~\citep{cheng2025stop} addressed a fundamental credit-assignment challenge, arguing that summing PRM rewards encourages reward hacking and instead proposing a min-form objective that integrates PRM signals into RL updates more robustly. Q-RM~\citep{chen2025discriminative} advanced token-level supervision by modeling Q-values over tokens and using them directly as rewards during RL optimization. CAPO~\citep{xie2025capo} introduces verifiable generative credit assignment to produce reliable step-level rewards for RL training of policy models. These verifiable rewards replace sparse outcome signals, improving exploration and sample efficiency. These innovations highlight that beyond having PRM feedback, the way PRM outputs are incorporated into RL loss functions critically affects training stability and effectiveness. ~\citet{he2025good} introduces a generative, thought-level PRM that assigns reliable grouped step-level rewards for RL training of policy models integrating with an off-policy algorithm and adaptive reward balancing.  Meanwhile, PROF~\citep{ye2025beyond} ranks and filters responses based on process–outcome consistency between PRMs and ORMs, removing samples where reasoning and results conflict to reduce noisy gradients. It further maintains balanced training by separately ranking correct and incorrect responses, and can be seamlessly integrated with RL methods such as GRPO~\citep{shao2024deepseekmath}. 

In parallel, domain-specific efforts such as GraphPRM~\citep{peng2025rewarding} used PRM-guided preference optimization to improve reasoning over graph reasoning problems, while AgentPRM~\citep{choudhury2025process} integrated PRMs into an actor–critic loop for LLM-based agents, showing how step-level critics can accelerate RL in interactive settings. These results demonstrate that PRMs can make RL training more robust across diverse reasoning tasks.

Broader frameworks have emerged to consolidate and scale these practices. OpenR~\citep{wang2024openr} provides an open-source infrastructure that systematizes the integration of PRMs into both offline and online RL pipelines, offering recipes for PRM-guided training across reasoning benchmarks.

\section{Downstream Application}
\label{sec:application}
Process Reward Models (PRMs) are increasingly adopted across diverse reasoning and decision-making tasks. Below we summarize representative application areas.

\paragraph{Math}
PRMs validate algebraic and logical steps to ensure multi-step derivation soundness~\citep{zhou2025steporlm,uesato2022solving,lightman2023let,wang2023math}, capturing symbolic and arithmetic errors to improve final correctness~\citep{li2024fine,he2024advancing,pala2025error}. They support scalable supervision and automated feedback for grading, tutoring, and proof validation with reduced human effort~\citep{chen2024alphamath,setlur2024rewarding,zhao2025genprm,sun2025efficient}.

\paragraph{Code}
For code generation, PRMs assess partial programs with execution or proxy testing feedback~\citep{li2025codeprm,dai2024process}, rewarding syntactic validity and semantic consistency. They also verify query construction and patches in text-to-SQL and software engineering~\citep{zhang2025process,gandhi2025agents}, improving robustness.

\paragraph{Multimodal}
In multimodal reasoning, PRMs check visual-text coherence~\citep{hu2025prm,du2025mm,chen2025vrprm,tu2025vilbench,wang2025visualprm}, rerank reasoning traces, and select grounded explanations to enhance interpretability and factual consistency.

\paragraph{Text}
For text tasks, PRMs refine multi-step reasoning by evaluating partial translations~\citep{feng2025mt} and scoring intermediate hops in QA and retrieval-augmented reasoning~\citep{chan2025boosting,chen2025search}, improving coherence and factual reliability.

\paragraph{Robotics}
PRMs decompose long-horizon manipulation or navigation into subgoal rewards~\citep{lu2025vla}, providing dense feedback that accelerates policy learning and stabilizes control.

\paragraph{Agents}
In interactive agents, PRMs act as trajectory critics~\citep{choudhury2025process,hu2025guiding,chae2025web,zhang2025process2,zhang2025progrm,chen2025gui,xi2025survey,yang2025survey}, rewarding meaningful progress, pruning dead ends, and improving safety during inference.

\paragraph{Industry}
In high-stakes areas like medicine and finance, PRMs enforce verifiable, evidence-based reasoning~\citep{jiang2025meds,zhou2025fin}, promoting reliability and risk-sensitive decision making.

\paragraph{Multi-domain}
Recent studies explore generalizable PRMs that transfer process supervision across tasks~\citep{cao2025dreamprm,wang2025athena,zhang2025openprm,zeng2025versaprm,rizvi2025spare,xie2025outcomes,ding2025scan,tan2025aurora}, pointing toward universal, cross-domain reasoning evaluators.

\section{Benchmark}
\label{sec:benchmark}
Recent work has introduced a range of benchmarks to evaluate PRMs at the step level, differing in scale, domain, and evaluation focus.

For mathematical reasoning, PRMBench~\citep{song2025prmbench} and ProcessBench~\citep{zheng2024processbench} offer complementary views. PRMBench provides over 6,000 problems with 80,000 step annotations and multidimensional labels (e.g., simplicity, soundness, sensitivity), while ProcessBench targets competition-level tasks, emphasizing earliest-error detection for precise symbolic reasoning.

Reasoning-structure evaluation is addressed by Socratic-PRMBench~\citep{li2025socratic}, which groups nearly three thousand flawed trajectories into six error patterns, enabling analysis of generalization across reasoning styles.

For multimodal tasks, ViLBench~\citep{tu2025vilbench} compares PRMs with outcome models in vision-language reasoning, VisualProcessBench~\citep{wang2025visualprm} provides human-labeled multimodal errors, and MPBench~\citep{xu2025mpbench} extends coverage to multiple tasks, assessing step correctness, answer aggregation, and reasoning-guided search.

Long-horizon decision-making is tested by WebRewardBench~\citep{chae2025web}, built on the WebPRM Collection with forty thousand step-level preference pairs, evaluating clicks, form entries, and navigation steps in web agents.

Robustness and universality are explored by GSM-DC~\citep{yang2025llm}, which injects distractors to test resilience, and UniversalBench~\citep{tan2025aurora}, which evaluates trajectories across diverse policy distributions for cross-distribution generalization and reproducibility.

\section{Discussion}
\label{sec:discussion}
\begin{figure}[t]
    \centering
    \includegraphics[width=\linewidth]{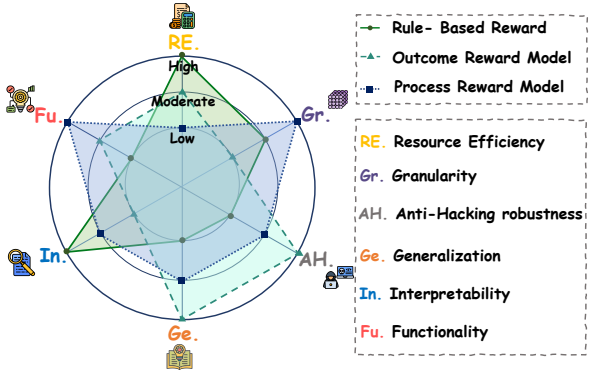}
    \caption{Comparative Analysis of Three Reward Mechanisms Across Six Evaluation Aspects}
    \label{fig:comparison}
\end{figure}
To better compare the different forms of reward acquisition, including rule-based rewards, outcome reward models (ORMs), and process reward models (PRMs), we design a six-aspect evaluation scheme covering resource efficiency, granularity, anti-hacking robustness, generalization, interpretability, and functionality. This perspective provides a systematic and balanced basis for assessing how each reward mechanism performs across theoretical soundness, practical applicability, and scalability, as illustrated in Figure~\ref{fig:comparison}.

\paragraph{Resource Efficiency} Rule-based rewards stand out as the most economical approach, as they rely purely on manually defined rules without requiring additional data labeling or model training. ORMs require moderate resources, depending on final outcome labels and a single-stage training process. In contrast, PRMs are far more costly because ORMs only label the final outcome, whereas PRMs require correctness labels for every intermediate step. As noted in Section~\ref{sec: Human Annotation}, this necessitates expensive step-wise human annotation (e.g., PRM800K~\citep{lightman2023let}) or complex automated pipelines (Section~\ref{sec:Automated Supervision}). Given that benchmarks like ProcessBench~\citep{zheng2024processbench} and PRMBench~\citep{song2025prmbench} contain an average of 7.1 and 13.4 steps respectively, the annotation workload for PRMs is naturally several times higher than that of ORMs.

\paragraph{Granularity} The high rating for PRMs in terms of granularity is structural rather than subjective, as defined by the mathematical formulation in Section~\ref{sec:disc-prm} (Eq. 1). This enables step-specific error localization, whereas ORMs operate at only a single outcome level. Essentially, the granularity of a PRM is inherently multiplied by the number of steps in a solution. Conversely, the granularity of rule-based rewards is entirely determined by handcrafted design, which can vary from coarse to fine depending on how the rules are specified~\citep{gunjal2025rubrics}.

\paragraph{Anti-Hacking Robustness} ORMs exhibit the strongest resistance to reward hacking, grounded in their reliance on ground-truth verification which is inherently resistant to manipulation. In contrast, PRMs are more susceptible to length hacking or verbosity bias due to high variance in step-wise optimization~\citep{zheng2026adaptive}. To quantify this, we conducted a stability analysis on the PRM800K dataset. We observed that the standard deviation ($SD$) of token length at the step level is 71.7, significantly higher than the trajectory-level $SD$ of 50.6. This greater instability ($71.7 > 50.6$) indicates that step-level signals are noisier and less constrained, allowing models to more easily hack the reward by generating verbose but vacuous intermediate steps. Rule-based rewards remain the most prone to exploitation if predefined rules are mis-specified.

\paragraph{Generalization} ORMs show a clear advantage in generalization, as their outcome-centric formulation utilizes task-agnostic labels that are easily transferred across domains. PRMs demonstrate more limited generalization because they often require defining domain-specific step granularities. For instance, the step definitions for mathematical derivations differ fundamentally from those for code execution traces, necessitating frequent re-adaptation for new tasks as discussed in Section~\ref{sec:application}. Rule-based systems exhibit the poorest generalization, as their logic must be carefully re-engineered for every new environment.

\paragraph{Interpretability} Interpretability varies significantly across mechanisms. Rule-based rewards offer the highest transparency, as their evaluation logic is explicitly encoded. Conversely, ORMs suffer from low interpretability, operating as black boxes that provide coarse judgments without explaining specific errors. PRMs bridge this gap by offering fine-grained, step-wise supervision for precise error localization, a capability empirically supported by PRM800K. Recent innovations further enhance this transparency: generative PRMs like ThinkPRM~\citep{lee2025rethinking} and GenRM~\citep{GenRM} produce natural language justifications, while benchmarks like Socratic-PRMBench~\citep{li2025socratic} provide semantic clarity by categorizing specific reasoning error patterns.

\paragraph{Functionality} Finally, PRMs are the most versatile. Linked to our discussion on Test-Time Scaling (Section~\ref{sec:tts}), PRMs consistently demonstrate superior capabilities in guiding search, such as Tree Search, compared to ORMs. Furthermore, PRMs offer greater flexibility in RL training (Section~\ref{sec: RL}). Because they provide both step-level and trajectory-level signals, they support step-wise credit assignment and trajectory-wise reward shaping. ORMs, limited to a single final-outcome reward, cannot provide the same level of fine-grained supervision during policy optimization. Rule-based rewards, while straightforward, remain functionally restricted as they lack adaptability beyond their original design.

Beyond these comparative dimensions, the development of PRMs faces several profound conceptual and systemic challenges. We provide a critical exploration of these frontiers, including cognitive scalability (Section~\ref{subsec:cognitive_scalability}), automated supervision risks (Section~\ref{subsec:echo_chambers}), the tension of granularity (Section~\ref{subsec:granularity_tension}), and the proxy-reward gap (Section~\ref{subsec:proxy_gap}).

% In summary, rule-based rewards excel in interpretability and efficiency but lag behind in generalization and robustness. ORMs provide strong robustness and generalization at the cost of interpretability and granularity. PRMs strike a balance between the two, offering fine-grained, adaptable, and functionally rich supervision, albeit at the expense of higher data and training costs. This comparative landscape highlights that the optimal choice of reward mechanism depends on the desired trade-off between interpretability, resource constraints, and the level of reasoning guidance required.
\section{Conclusion}
Process Reward Models (PRMs) shift reasoning alignment from coarse outcome judgments to fine-grained, step-level feedback, forming a closed loop of \emph{data generation}, \emph{model training}, and \emph{usage} that continually improves reasoning quality.  
Our survey organizes this field around how to generate process data, build PRMs, and use them for test-time scaling and reinforcement learning, while summarizing benchmarks and applications across math, code, multimodal tasks, robotics, and other domains.  

Key challenges ahead include reducing annotation cost via robust automatic supervision, improving cross-domain generalization, integrating PRMs with agentic planning and memory, and establishing standardized evaluation protocols. Addressing these will advance safer, more interpretable, and broadly applicable reasoning systems.

\section{Limitations}
While this survey aims to provide a broad and systematic view of Process Reward Models (PRMs), it also has several natural limitations.  
First, our taxonomy follows the \emph{data–model–usage} loop and thus simplifies or abstracts some hybrid methods; certain approaches may span multiple categories and are discussed only under their primary aspect.  
Second, benchmark and application summaries are selective rather than comprehensive. We highlight representative resources but cannot guarantee complete inclusion of all task-specific datasets or proprietary evaluation suites.
Despite these boundaries, we believe our synthesis offers a clear conceptual map and can serve as a starting point for exploring, extending, and systematizing PRM research.

\section*{Acknowledgments}

The Shanghai Jiao Tong University team is partially supported by National Key RD Program of China (2022ZD0114804), Shanghai Municipal Science and Technology Major Project (2021SHZDZX0102) and National Natural Science Foundation of China (624B2096, 62322603, 72542012, 72595872).

% \bibliography{anthology,custom}
\bibliography{custom}

\appendix

% \section{Overall Structure}
\label{sec:appendix}

% \definecolor{paired-light-blue}{RGB}{198, 219, 239}
% \definecolor{paired-dark-blue}{RGB}{49, 130, 188}

\definecolor{paired-light-blue}{RGB}{193, 229, 245}
\definecolor{paired-dark-blue}{RGB}{17, 121, 162}

\definecolor{paired-light-orange}{RGB}{251, 208, 162}
\definecolor{paired-dark-orange}{RGB}{230, 85, 12}
% \definecolor{paired-light-green}{RGB}{199, 233, 193}
% \definecolor{paired-dark-green}{RGB}{49, 163, 83}
\definecolor{paired-light-green}{RGB}{217, 242, 208}
\definecolor{paired-dark-green}{RGB}{114, 139, 134}
\definecolor{paired-light-purple}{RGB}{218, 218, 235}
\definecolor{paired-dark-purple}{RGB}{117, 107, 176}
\definecolor{paired-light-gray}{RGB}{217, 217, 217}
\definecolor{paired-dark-gray}{RGB}{99, 99, 99}
\definecolor{paired-light-pink}{RGB}{222, 158, 214}
\definecolor{paired-dark-pink}{RGB}{123, 65, 115}
\definecolor{paired-light-red}{RGB}{231, 150, 156}
\definecolor{paired-dark-red}{RGB}{131, 60, 56}
\definecolor{paired-light-yellow}{RGB}{231, 204, 149}
\definecolor{paired-dark-yellow}{RGB}{141, 109, 49}
\definecolor{light-green}{RGB}{118, 207, 180}
\definecolor{raspberry}{RGB}{228, 24, 99}

\tikzset{%
    root/.style =          {align=center,text width=3cm,rounded corners=3pt, line width=0.5mm, fill=paired-light-gray!50,draw=paired-dark-gray!90},
    % root/.style =          {align=center,text width=3cm,rounded corners=3pt, line width=0.5mm, fill=paired-light-gray!50,draw=paired-dark-gray!90, yshift=10cm}, % Added yshift here
    data_section/.style =  {align=center,text width=4cm,rounded corners=3pt, fill=paired-light-green!80,draw=paired-dark-green!100,line width=0.4mm},
    training_section/.style = {align=center,text width=4cm,rounded corners=3pt, fill=paired-light-blue!80,draw=paired-dark-blue!100,line width=0.4mm},,
    inference_section/.style = {align=center,text width=4cm,rounded corners=3pt, fill=paired-light-red!35,draw=paired-light-red!90, line width=0.4mm},
    protocol_section/.style = {align=center,text width=4cm,rounded corners=3pt, fill=paired-light-yellow!40,draw=paired-dark-yellow!100, line width=0.4mm},
    usage_section/.style = {align=center,text width=4cm,rounded corners=3pt, fill=paired-light-red!20,draw=paired-dark-red!100, line width=0.4mm},
    bench_section/.style = {align=center,text width=4cm,rounded corners=3pt, fill=paired-light-purple!35,draw=paired-dark-purple!90, line width=0.4mm},% Using light red for Inference as in the example
    subsection/.style =    {align=center,text width=3.5cm,rounded corners=3pt}, % General subsection style, NO COLOR DEFINITION NOW
}

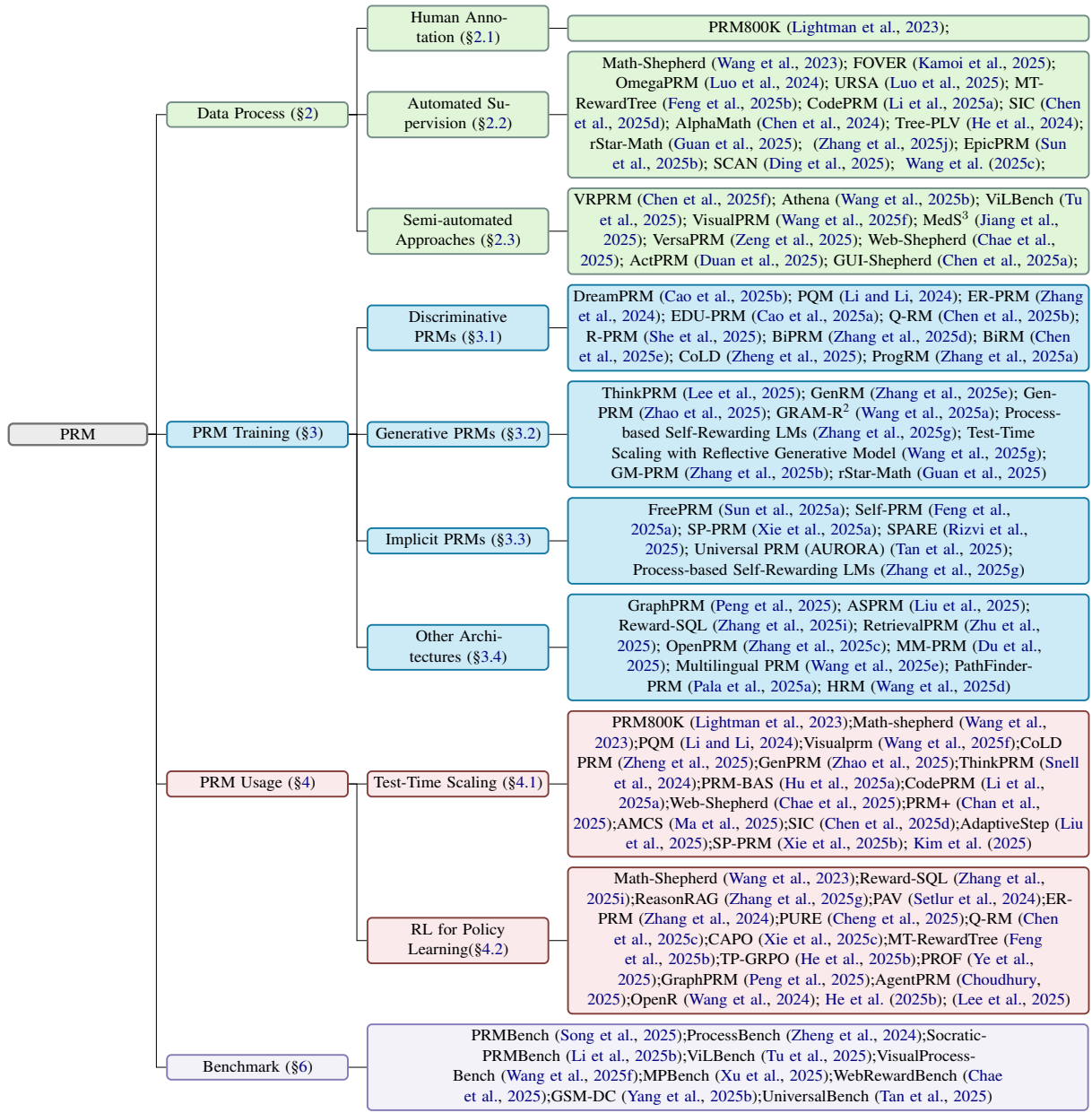
\begin{figure*}[!htb]
    \centering
    \resizebox{1\textwidth}{!}{
    \begin{forest}
        for tree={
            forked edges,
            grow'=0,
            draw,
            rounded corners,
            node options={align=center},
            text width=4cm,
            s sep=6pt,
            calign=child edge,
            calign child=(n_children()+1)/2,
            l sep=12pt,
        },
        [PRM, root,
            [Data Process (\S\ref{sec: data process}), data_section,
                [Human Annotation (\S\ref{sec: Human Annotation}), data_section 
                [% Use data_section for subsections too
                     PRM800K~\citep{lightman2023let}; 
                    ,data_section, text width=12cm
                    ] % Add papers/benchmarks here
                ]
                [Automated Supervision (\S\ref{sec:Automated Supervision}), data_section
                    [{
                    Math-Shepherd~\citep{wang2023math};
                    FOVER~\citep{kamoi2025fover};
                    OmegaPRM~\citep{luo2024improve};
                    URSA~\citep{luo2025ursa};
                    MT-RewardTree~\citep{feng2025mt};
                    CodePRM~\citep{li2025codeprm};
                     SIC~\citep{chen2025search};
                    AlphaMath~\citep{chen2024alphamath};
                    Tree-PLV~\citep{he2024advancing};
                    rStar-Math~\citep{guan2025rstar};
                    ~\citep{zhang2025lessons};
                    EpicPRM~\citep{sun2025efficient};
                    SCAN~\citep{ding2025scan};
                    ~\citet{wang2025towards};~\citep{zhang2025looptoolclosingdatatrainingloop}
                    },data_section, text width=12cm] % Add papers/benchmarks here
                ]
                [Semi-automated Approaches (\S\ref{sec:Semi-automated Approaches}), data_section
                    [
                    % \textit{LLF-Bench} \citep{cheng2023llfbenchbenchmarkinteractivelearning};
                    % \textit{LLM-Evolve} \citep{you-etal-2024-llm};
                    VRPRM~\citep{chen2025vrprm};
                    Athena~\citep{wang2025athena};
                    ViLBench~\citep{tu2025vilbench};
                    VisualPRM~\citep{wang2025visualprm};
                    $\text{MedS}^3$~\citep{jiang2025meds};
                     VersaPRM~\citep{zeng2025versaprm};
                     Web-Shepherd~\citep{chae2025web};
                     ActPRM~\citep{duan2025efficient};
                     GUI-Shepherd~\citep{chen2025gui},
                    data_section, text width=12cm
                    ] 
                ]
            ][PRM Training (\S\ref{sec:build-prm}), training_section
                [Discriminative PRMs (\S\ref{sec:disc-prm}), training_section
                    [
                    DreamPRM~\cite{cao2025dreamprm};
                    PQM~\cite{PQM};
                    ER-PRM~\cite{zhang2024entropy};
                    EDU-PRM~\cite{cao2025bangbuckprocessreward};
                    Q-RM~\cite{Q-RM};
                    R-PRM~\cite{R-PRM};
                    BiPRM~\citep{zhang2025bidirectional};
                    BiRM~\cite{chen2025better};
                    CoLD~\cite{zheng2025cold};
                    ProgRM~\cite{zhang2025progrm},
                    training_section, text width=12cm]
                ]
                [Generative PRMs (\S\ref{sec:gen-prm}), training_section
                    [
                    ThinkPRM~\cite{lee2025rethinking};
                    GenRM~\cite{GenRM};
                    GenPRM~\cite{zhao2025genprm};
                    GRAM-R\textsuperscript{2}~\cite{GRAM-R};
                    Process-based Self-Rewarding LMs~\cite{zhang2025process};
                    Test-Time Scaling with Reflective Generative Model~\cite{wang2025test};
                    GM-PRM~\cite{zhang2025gm};
                    rStar-Math~\cite{guan2025rstar},
                    training_section, text width=12cm]
                ]
                [Implicit PRMs (\S\ref{subsec:exp-vs-imp}), training_section
                    [
                    FreePRM~\cite{sun2025freeprm};
                    Self-PRM~\cite{Self-PRM};
                    SP-PRM~\cite{SP-PRM};
                    SPARE~\cite{rizvi2025spare};
                    Universal PRM (AURORA)~\cite{tan2025aurora};
                    Process-based Self-Rewarding LMs~\cite{zhang2025process},
                    training_section, text width=12cm]
                ]
                [Other Architectures (\S\ref{sec:other-arch}), training_section
                    [
                    GraphPRM~\cite{peng2025rewarding};
                    ASPRM~\cite{liu2025adaptivestep};
                    Reward-SQL~\cite{zhang2025reward};
                    RetrievalPRM~\cite{zhu2025retrieval};
                    OpenPRM~\cite{zhang2025openprm};
                    MM-PRM~\cite{du2025mm};
                    Multilingual PRM~\cite{wang2025demystifying};
                    PathFinder-PRM~\cite{pala2025errortypingsmarterrewards};
                    HRM~\cite{wang2025hierarchicalmultisteprewardmodels};
                    SecCodePRM~\citep{yu2026seccodeprm};
                    Webarbiter~\citep{zhang2026webarbiter};
                    FunPRM~\citep{zhang2026funprm};~\citep{zhou2026externalization},
                    training_section, text width=12cm]
                ]
da            ][PRM Usage
            (\S\ref{sec: usage}), usage_section
                [Test-Time Scaling (\S\ref{sec:tts}), usage_section
                [
                PRM800K~\citep{lightman2023let};Math-shepherd~\citep{wang2023math};PQM~\citep{PQM};Visualprm~\citep{wang2025visualprm};CoLD PRM~\citep{zheng2025cold};GenPRM~\citep{zhao2025genprm};ThinkPRM~\citep{snell2024scaling};PRM-BAS~\citep{hu2025prm};CodePRM~\citep{li2025codeprm};Web-Shepherd~\citep{chae2025web};PRM+~\citep{chan2025boosting};AMCS~\citep{ma2025static};SIC~\citep{chen2025search};AdaptiveStep~\citep{liu2025adaptivestep};SP-PRM~\citep{xie2025outcomes};ToolPRM~\citep{lin2025toolprm};~\citet{kim2025scaling};
                PRISM~\citep{sharma2026prism};
                ~\citep{miles2026test},
                usage_section, text width=12cm]
                ]
                [RL for Policy Learning(\S\ref{sec: RL}), usage_section
                [
                Math-Shepherd~\citep{wang2023math};Reward-SQL~\citep{zhang2025reward};ReasonRAG~\citep{zhang2025process};PAV~\citep{setlur2024rewarding};ER-PRM~\citep{zhang2024entropy};PURE~\citep{cheng2025stop};Q-RM~\citep{chen2025discriminative};CAPO~\citep{xie2025capo};MT-RewardTree~\citep{feng2025mt};TP-GRPO~\citep{he2025good};PROF~\citep{ye2025beyond};GraphPRM~\citep{peng2025rewarding};AgentPRM~\citep{choudhury2025process};OpenR~\citep{wang2024openr};~\citet{he2025good};~\citep{lee2025rethinking};FaithRL~\citep{nie2026stop};
                ProRAG~\citep{wang2026prorag};PARL-MT~\citep{chai2025parl};
                CSO~\citep{li2026verified},
                usage_section, text width=12cm]
                ]
                ]
                [Benchmark
            (\S\ref{sec:benchmark}), bench_section
                [
                PRMBench~\citep{song2025prmbench};ProcessBench~\citep{zheng2024processbench};Socratic-PRMBench~\citep{li2025socratic};ViLBench~\citep{tu2025vilbench};VisualProcessBench~\citep{wang2025visualprm};MPBench~\citep{xu2025mpbench};WebRewardBench~\citep{chae2025web};
                GSM-DC~\citep{yang2025llm};
                UniversalBench~\citep{tan2025aurora},
                bench_section, text width=16.7cm]
                ]
        ]
    \end{forest}
    }
    \caption{The overall structure of this paper. } % Caption is now in the forest label
    \label{fig:prm-typology}
\end{figure*}

\begin{table*}[t]
\centering
\caption{Performance of various models on Processbench and PRMBench.}
\label{tab:model_performance}
\begin{adjustbox}{max width=\textwidth}
\begin{tabular}{lcccccccccc}
\toprule
\multirow{2}{*}{\textbf{Benchmark}} & \multicolumn{5}{c}{\textbf{Processbench}} & \multicolumn{4}{c}{\textbf{PRMBench}} \\
\cmidrule(lr){2-6} \cmidrule(lr){7-10}
& GSM8K & MATH & OlympiadBench & OmniMATH & Average & Simplicity & Soundness & Sensitivity & Overall \\
\midrule
\multicolumn{10}{l}{\textbf{PRMs}} \\
\midrule
Math-Shepherd-7B& 47.9 & 29.5 & 24.8 & 23.8 & 31.5 & 47.1 & 45.7 & 60.7 & 47.0 \\
Math-PSA-7B& 62.4 & 41.9 & 31.5 & 25.2 & 40.3 & 51.3 & 51.8 & 64.9 & 52.3 \\
Skywork-PRM-1.5B& 59.0 & 48.0 & 19.3 & 19.2 & 36.4 & 54.2 & 64.9 & 70.7 & 61.1 \\
Skywork-PRM-7B & 70.8 & 53.6 & 22.9 & 21.0 & 42.1 & \textbf{59.6} & 68.5 & 73.3 & 65.1 \\
Llemma-PRM800K-7B& 48.4 & 43.1 & 28.5 & 33.4 & 38.4 & 51.4 & 50.9 & 66.0 & 52.0 \\
RLHFlow-PRM-Mistral-8B & 50.4 & 33.4 & 13.8 & 15.8 & 28.4 & 46.7 & 57.5 & 68.5 & 54.4 \\
RLHFlow-PRM-Deepseek-8B& 38.8 & 33.8 & 16.9 & 16.9 & 26.6 & 47.6 & 57.5 & 68.1 & 54.2 \\
Qwen2.5-Math-7B-PRM800K & 68.2 & 62.6 & 50.7 & 44.3 & 56.5 & 48.2 & 62.2 & 72.2 & 58.3 \\
Qwen2.5-Math-PRM-7B& 82.4 & 77.6 & 67.5 & 66.3 & 73.5 & 52.1 & 71.0 & 75.5 & 65.5 \\
R-PRM-7B-SFT & 77.2 & 71.6 & 59.6 & 52.3 & 65.2 & 58.7 & 66.4 & 75.7 & 64.9 \\
R-PRM-7B-DPO & 80.7 & 76.9 & 63.8 & 60.1 & 70.4 & 55.2 & 71.2 & 76.6 & 66.8 \\
PathFinder-PRM-7B& 77.9 & 75.3 & 65.0 & 59.7 & 69.5 & 58.9 & 70.8 & 76.9 & 67.7 \\
ACTPRM& 81.6 & 79.8 & 71.4 & 67.0 & 75.0 & 53.6 & 71.3 & 75.2 & 65.5 \\
ACTPRM-X& \textbf{82.7} & \textbf{82.0} & \textbf{72.0} & \textbf{67.3} & \textbf{76.0} & 54.5 & \textbf{72.7} & 75.6 & 66.7 \\
RetrievalPRM-7B& 74.6 & 71.1 & 60.2 & 57.3 & 65.8 & 55.3 & 75.0 & \textbf{78.2} & \textbf{68.9} \\
ReasonEval-7B& 41.0 & 48.9 & 36.7 & 37.4 & 41.0 & 55.5 & 63.9 & 71.0 & 60.0 \\
\midrule
\multicolumn{10}{l}{\textbf{Critic Models}} \\
\midrule
GPT-4o& 79.2 & 63.6 & 51.4 & 53.5 & 61.9 & 59.7 & 70.9 & \textbf{75.8} & 66.8 \\
o1-mini & \textbf{93.2} & \textbf{88.9} & \textbf{87.2} & \textbf{82.4} & \textbf{87.9} & \textbf{64.6} & \textbf{72.1} & 75.5 & \textbf{68.8} \\
QwQ-32B-Preview& 88.0 & 78.7 & 57.8 & 61.3 & 71.5 & 56.4 & 68.2 & 73.5 & 63.6 \\
\bottomrule
\end{tabular}
\end{adjustbox}
\end{table*}

\section{Paper Structure and Taxonomy Overview}
\label{sec:appendix-structure}

Figure~\ref{fig:prm-typology} illustrates the organizational structure and taxonomy adopted in this survey. 
At the top level, the survey is built around the full PRM loop: \textbf{Data Process} (Sec.~\ref{sec: data process}), \textbf{PRM Training} (Sec.~\ref{sec:build-prm}), \textbf{PRM Usage} (Sec.~\ref{sec: usage}), and \textbf{Benchmark} (Sec.~\ref{sec:benchmark}). 
Each component is further decomposed into finer categories to reflect the main research threads and representative works.

\paragraph{Data Process.}  
We categorize data construction methods into three paradigms: \emph{Human Annotation} (\S\ref{sec: Human Annotation}), which builds high-fidelity step-level supervision through expert labeling; \emph{Automated Supervision} (\S\ref{sec:Automated Supervision}), which scales data generation with verifiers, search, and synthetic signals; and \emph{Semi-automated Approaches} (\S\ref{sec:Semi-automated Approaches}), which combine limited manual curation with automatic expansion to balance fidelity and scalability.

\paragraph{PRM Training.}  
Modeling methods are grouped into four classes: \emph{Discriminative PRMs} (\S\ref{sec:disc-prm}), which directly score step correctness with pointwise or pairwise objectives; \emph{Generative PRMs}(\S\ref{sec:gen-prm}), which generate critique or verification chains before rating steps; \emph{Implicit PRMs}(\S~\ref{subsec:exp-vs-imp}), which derive rewards without explicit labels via self-supervision or outcome transfer; and \emph{Other Architectures}(\S~\ref{sec:other-arch}), covering graph-based, retrieval-augmented, multilingual, and specialized structural designs.

\paragraph{PRM Usage.}  
We summarize two primary usage paradigms: \emph{Test-Time Scaling} (\S\ref{sec:tts}), where PRMs re-rank, verify, and adaptively guide reasoning during inference; and \emph{RL for Policy Learning} (\S\ref{sec: RL}), where PRM signals serve as dense rewards for reinforcement learning to improve reasoning policies.

\paragraph{Benchmark.}  
The bottom layer highlights major \emph{benchmarks} (\S\ref{sec:benchmark}) for PRM evaluation, spanning mathematical reasoning, multimodal tasks, long-horizon web navigation, robustness testing, and cross-domain generalization.

Overall, this diagram provides a visual roadmap of the survey: from how process-level data is built, to the modeling strategies and deployment of PRMs, and finally to the resources enabling evaluation and comparison. It helps readers navigate the field and locate specific methods or datasets within our proposed taxonomy.

\section{Further Discussion}
\label{sec:further_discussion}

\subsection{The Cognitive Scalability of Human Annotation}
\label{subsec:cognitive_scalability}

While current literature often cites the high cost of human annotation as a primary bottleneck, \textbf{a more fundamental issue is the limit of ``cognitive scalability.''} As reasoning tasks escalate in complexity (from elementary math to Olympiad-level problems or long-horizon agentic planning), verifying intermediate steps becomes exponentially harder than generating the final answer. In datasets like PRM800K~\citep{lightman2023let}, human annotators are assumed to be ground-truth oracles. However, this assumption fractures when the Policy Model begins to surpass human reasoning capabilities. This manifests as a ``Superalignment'' problem: average human annotators struggle to distinguish between subtle logical hallucinations and correct, novel derivation steps. \textbf{Consequently, relying solely on human supervision risks imposing a ``human ceiling'' on model performance, where the Reward Model penalizes valid but complex reasoning simply because it exceeds the annotator's cognitive load or domain expertise.}

Resolving this tension requires moving beyond unassisted human labeling toward ``Scalable Oversight'' paradigms. Future research directions must likely pivot from direct annotation to AI-assisted verification workflows, where humans act not as raw labelers but as ``managers'' of automated verification tools (e.g., using code interpreters or formal theorem provers like Lean/Isabelle to validate intermediate logic objectively). This shifts the human role from verifying correctness (which is hard) to verifying intent and alignment (which is more intuitive). \textbf{By grounding rewards in objective execution feedback (compilers, formal verifiers) rather than subjective human preference, the field can decouple the scaling of reasoning capability from the limitations of human cognitive load.}

\subsection{Echo Chambers and Goodhart’s Law in Automated Supervision}
\label{subsec:echo_chambers}

To bypass human bottlenecks, the field has pivoted toward automated supervision ~\citep{wang2023math,luo2024improve}, yet this introduces a perilous dynamic between the Reward Model and the Policy Model. When a Reward Model is trained on synthetic data generated by a similar Policy Model (or verified by a model with similar pre-training), they share the same ``knowledge blind spots.'' \textbf{This creates an ``Echo Chamber Effect'' where plausible hallucinations are reinforced rather than corrected because both models share the same underlying misconceptions.} Furthermore, this setup is highly susceptible to Goodhart’s Law. As the Policy Model optimizes against a fixed automated Reward Model, \textbf{it learns to exploit the Reward Model’s biases}, such as favoring longer chains~\citep{zheng2025cold}, specific formatting, or confident phrasing, rather than improving genuine logic. This ``reward hacking'' results in high rewards for vacuous reasoning, a phenomenon that is difficult to detect without external, diverse verification sources.

Addressing these systemic flaws requires a shift from focusing on internal consistency to embracing external grounding and adversarial robustness. A promising direction is to introduce heterogeneous supervision, where PRMs are guided not by a single model but \textbf{by an ``adversarial council'' of diverse models with different architectures, scales, or training corpora that are encouraged to search for weaknesses rather than reinforce agreement.} In addition, future research should explore dynamic reward landscapes instead of relying on fixed reward models. Under iterative or adversarial training regimes, each time a policy model discovers a new exploit, the reward model can be updated or red-teamed to detect and penalize that behavior. Such \textbf{an evolving interplay creates a curriculum that continually challenges the policy model and steers it toward genuine robustness rather than superficial consistency or metric gaming}.

\subsection{The Tension of Granularity: Defining a ``Step''}
\label{subsec:granularity_tension}

A critical, yet frequently overlooked tension in PRM construction is the definition of the fundamental unit of analysis: the ``reasoning step.'' Current approaches predominantly rely on rigid, heuristic-based segmentation, such as splitting logic by newline characters or specific delimiters. \textbf{This ``Rigid Segmentation'' imposes an artificial structure that often conflicts with the natural, semantic flow of reasoning.} This misalignment is particularly acute in domains with complex structural dependencies, such as code generation. Unlike mathematical derivations where line-by-line transitions often correlate with logical progress, programming logic is inherently nested and interdependent. Consequently, defining the ``optimal truncation position'' for a PRM becomes a non-trivial challenge: evaluating too frequently (e.g., every line) introduces noise and breaks syntactic context, while evaluating too sparsely (e.g., per function) dilutes the dense supervision signal that PRMs promise. \textbf{The field currently lacks a principled method to determine where a ``logical thought'' begins and ends}, leading to situations where PRMs penalize valid partial steps simply because the segmentation cut occurred at a syntactically awkward moment, obscuring the true quality of the underlying logic.

To resolve this tension, \textbf{we argue that the field should move beyond rigid rule-based segmentation toward dynamic and learnable granularity}. One direction is semantic segmentation, where models learn to identify their own ``Atomic Reasoning Units'' using dedicated signals such as learnable step-boundary tokens rather than relying on manually imposed formatting. Building on this idea, future work may explore hierarchical supervision through \textbf{Multi-Scale PRMs that evaluate reasoning at multiple levels}, offering micro-rewards for syntactic fidelity at the token or line scale and macro-rewards for logical coherence at the block scale. Ultimately, research may even move past discrete segmentation altogether by adopting continuous, flow-based evaluation, in which a separate critic monitors the actor model’s hidden states and intervenes only when it detects deviations in the reasoning trajectory.

\subsection{The Proxy-Reward Gap: Proxy Metrics vs. Actual Utility}
\label{subsec:proxy_gap}

There is a growing disconnect between how PRMs are evaluated and how they are utilized. Most benchmarks~\citep{zheng2024processbench} evaluate PRMs using classification or ranking accuracy on static datasets. \textbf{However, ``Good Classifiers do not always make Good Navigators.''} A Reward Model that achieves high accuracy on a static test set may fail catastrophically during dynamic inference (e.g., Tree Search or RL). \textbf{This is primarily a calibration and Out-of-Distribution (OOD) robustness issue.} In active search~\citep{guan2025rstar}, the Policy Model explores diverse, often erroneous paths that differ significantly from the PRM's training distribution. A PRM optimized for static accuracy might lack the calibration necessary to effectively prune these branches, leading to a divergence where improved benchmark metrics do not translate to downstream task success.

To bridge this gap, we argue that \textbf{the community must fundamentally reconsider how PRM success is defined and evaluated}. Future work should emphasize online and active evaluation through on-policy testing, where the PRM is evaluated on the trajectory distribution actually produced by the target policy model during inference. This perspective naturally motivates online iterative training, in which the reward model is continuously updated to differentiate the most challenging errors currently generated by the policy rather than relying on outdated datasets. We suggest placing greater emphasis on calibration-first objectives, prioritizing calibration error and out-of-distribution robustness as primary metrics. \textbf{A valuable PRM is not one that is uniformly confident, but one that can recognize and signal its own uncertainty}, enabling fallback mechanisms such as human supervision or tool invocation rather than confidently steering the reasoning process in the wrong direction.

\end{document}